\documentclass[conference]{IEEEtran}
\usepackage{booktabs}

\usepackage[listings,skins,breakable]{tcolorbox}
\usepackage{colortbl}
\usepackage{pgfplots}
\usepackage{pgfplotstable}
\usepackage{hyperref}
\usepackage{booktabs}
\usepackage{pdfpages}
\usepackage{multirow}
\usepackage[utf8]{inputenc}
\usepackage{authblk}
\usepackage[T1]{fontenc}
\usepackage{amsmath,amssymb}
\usepackage{paralist}
\usepackage{latexsym}
\usepackage{xcolor}
\usepackage{wasysym}
\usepackage{graphicx}
\usepackage{MnSymbol}
\usepackage{stackengine}
\usepackage{mdframed}
\usepackage{comment}
\usepackage{array}
\usepackage{lipsum}
\usepackage{algorithmic}
\usepackage{enumitem}
\usepackage{menukeys}
\renewcommand{\figurename}{Fig.}
\renewcommand{\tablename}{Tab.}

\pgfplotsset{
    caseStudy plot/.style={
    yticklabel=\pgfmathprintnumber{\tick}\,$\%$,
    ymin=0,
    ymax=100.01, 
    label style={font=\scriptsize},
                    tick label style={font=\scriptsize},
    visualization depends on={y \as \originalvalue},
  },
  percentage plot/.style={
    point meta=explicit,
    every node near coord/.append style={
    font=\scriptsize,
       color=black,
    }}}

\makeatletter
\def\ps@IEEEtitlepagestyle{%
  \def\@oddfoot{\mycopyrightnotice}%
  \def\@evenfoot{}%
}
\def\mycopyrightnotice{%
  {\begin{minipage}{\textwidth}
  \footnotesize \copyright 2021 IEEE. Personal use of this material is permitted. Permission from IEEE must be obtained for all other uses, in any current or future media, including reprinting\slash republishing this material for advertising or promotional purposes, creating new collective works, for resale or redistribution to servers or lists, or reuse of any copyrighted component of this work in other works.
  \end{minipage}
  }
  \gdef\mycopyrightnotice{}
}

\begin{document}

\title{Transfer Learning for Mining Feature Requests and Bug Reports from Tweets and App Store Reviews}

\author[1]{Pablo Restrepo Henao}
\author[2]{Jannik Fischbach}
\author[2]{Dominik Spies}
\author[3]{\\ Julian Frattini}
\author[4]{Andreas Vogelsang}

\affil[1]{\footnotesize Technical University of Munich, Germany, 
pablo.restrepo@tum.de}
\affil[2]{\footnotesize Qualicen GmbH, Germany, \{firstname.lastname\}@qualicen.de}
\affil[3]{\footnotesize Blekinge Institute of Technology, Sweden, julian.frattini@bth.se}
\affil[4]{\footnotesize University of Cologne, Germany, vogelsang@cs.uni-koeln.de}

\maketitle

\begin{abstract}
Identifying feature requests and bug reports in user comments holds great potential for development teams. However, automated mining of RE-related information from social media and app stores is challenging since (1) about 70\% of user comments contain noisy, irrelevant information, (2) the amount of user comments grows daily making manual analysis unfeasible, and (3) user comments are written in different languages. Existing approaches build on traditional machine learning (ML) and deep learning (DL), but fail to detect feature requests and bug reports with high Recall and acceptable Precision which is necessary for this task. In this paper, we investigate the potential of transfer learning (TL) for the classification of user comments. Specifically, we train both monolingual and multilingual \emph{BERT} models and compare the performance with state-of-the-art methods. We found that monolingual \emph{BERT} models outperform existing baseline methods in the classification of English App Reviews as well as English and Italian Tweets. However, we also observed that the application of heavyweight TL models does not necessarily lead to better performance. In fact, our multilingual \emph{BERT} models perform worse than traditional ML methods. 
\end{abstract}

\section{Introduction}
\textbf{Motivation} Social media and app stores allow users to communicate their opinions about apps. Studies have shown that user comments can contain valuable information for development teams, such as feature requests and problem reports~\cite{palomba15}. However, a manual analysis of user comments is cumbersome, as the number of reviews grows rapidly every day. For example, popular apps like Facebook receive around 4,000 reviews daily~\cite{maalej13}. About 70\% of these user comments contain noisy, irrelevant information while 30\% actually express requirements related to the app~\cite{maalej13,guzman16}. As a result, development teams face a two-fold challenge: First, they need to manage the large volume of user comments that are not necessarily written in English due to the global distribution of modern applications. Second, they need to filter out only those comments that are relevant for further app development.

\textbf{Research Gap} Automated mining of RE-related information from social media and app stores is an active research field. Several studies utilize traditional machine learning (ML) approaches to identify feature requests and bug reports in user comments~\cite{Guzman15,Maalej2017,williams17,panichella15}. However, their performance is highly dependent on handcrafted features. Stanik et al.~\cite{stanik19} propose to use deep learning (DL) approaches, which are able to automatically extract useful features from raw text. Nevertheless, the use of DL approaches did not lead to a significant performance gain in contrast to traditional ML approaches, since the existing approaches still fail to extract user comments with a performance that allows their use in practice. For example, Stanik et al.~\cite{stanik19} achieve a poor Precision score of 0.51 for problem reports and 0.40 for feature requests when categorizing English tweets. This causes development teams to be overwhelmed with a high amount of False Positives. Consequently, the existing approaches solve the aforementioned challenge only to a limited extent, since users still have to sort out many false predictions manually, which negatively affects user acceptance~\cite{Femmer2018}.

\textbf{Objective} Due to the scarcity of suitable training data, we currently observe a shift towards the usage of transfer learning (TL). TL models are pre-trained on large linguistic corpora and can be fine-tuned for a specific task on smaller amounts of supervised data. Recent studies demonstrate that TL models outperform existing approaches in a number of tasks. Hey et al.~\cite{Hey20} employ the Bidirectional Encoder Representations from Transformers (BERT) model~\cite{devlin19} to classify requirements into functional / non-functional and achieves new state-of-the-art performance. Fischbach et al.~\cite{fischbach21} show that \emph{BERT} can also help to detect causal relationships in requirements. So far, however, we lack knowledge about the potential of TL methods for the classification of user comments into problem reports and feature requests. Surprisingly, the RE community has not yet studied whether the cross-lingual generalization ability of multilingual TL models is useful to classify user comments written in different languages. We address this research gap and investigate the following research question: \textit{Does the application of monolingual/multilingual BERT models lead to new state-of-the-art results in the classification of problem reports and feature requests from user comments?}

\textbf{Contributions} In this paper, we make the following contributions (C):

\begin{enumerate}[label=\bfseries C\arabic*:,leftmargin=*,labelindent=0em]
    \item We demonstrate that monolingual \emph{BERT} models achieve new state-of-art results in the classification of English App Reviews as well as English and Italian Tweets.  
    \item We prove that the \emph{BERT} model does not necessarily lead to a performance gain. In our case, the multilingual \emph{BERT} models even perform significantly worse than the existing baseline systems.
    \item To strengthen transparency and facilitate replication, we disclose our code.\footnote{\label{note1}Our code can be found at \url{https://github.com/prestrepoh/Topic-Mining-From-Tweets-and-App-Reviews}.} We also provide a Jupyter notebook that can be used to test our trained models on previously unseen data. We invite fellow researchers and practitioners to utilize our online demo for the classification of their user comments.\footnote{\label{note2}Our online demo can be accessed at \url{https://colab.research.google.com/drive/1y5gsiFEWtQrsXy7uyu-KUPq_xWJnRSH2?usp=sharing}.}
\end{enumerate}

\section{Research Methodology}
We are interested in the potential of \emph{BERT} for the classification of user comments into problem reports and feature requests. Specifically, we want to determine whether \emph{BERT} is capable of achieving new state-of-the-art results. For this purpose, we compare its performance with the approaches developed by Stanik et al.~\cite{stanik19}, which represent the hitherto best-performing ML and DL models for this task. We use the data and annotations of Maalej et al.~\cite{Maalej2017} and Stanik et al.~\cite{stanik19} as our benchmark data sets. We decided to build on these data sets since: (1) they were carefully annotated using peer, manual
content analysis and contain only user comments for which the annotators agreed on the same label. (2) they serve as a gold standard corpus and allow us to compare our methods with existing state-of-the-art methods. In the following, we give an overview of the used data set and an insight into our evaluation procedure. 

\subsection{Study Objects \& Labels}\label{study_objects}
\begin{figure}
\centering
\begin{tikzpicture}
\begin{axis}[legend style={nodes={scale=0.6, transform shape}},
axis on top,
nodes near coords,
ylabel=Ratio of User Comments,
caseStudy plot,
ybar=5pt,
enlarge x limits={abs=1.2cm},
symbolic x coords={App Reviews EN, Tweet EN, Tweet IT},
xtick=data
]
\addplot coordinates {(App Reviews EN,22.4) (Tweet EN,28.2) (Tweet IT,21.6)};
\addplot coordinates {(App Reviews EN,17.1) (Tweet EN,13.5) (Tweet IT,16.4)};
\addplot coordinates {(App Reviews EN,60.4) (Tweet EN,58.1) (Tweet IT,61.9)};
\legend{Problem Report,Feature Request,Irrelevant}
\end{axis}
\end{tikzpicture}
\caption{Overview of Benchmark Data Sets}
\label{fig:data_sets}
\end{figure}
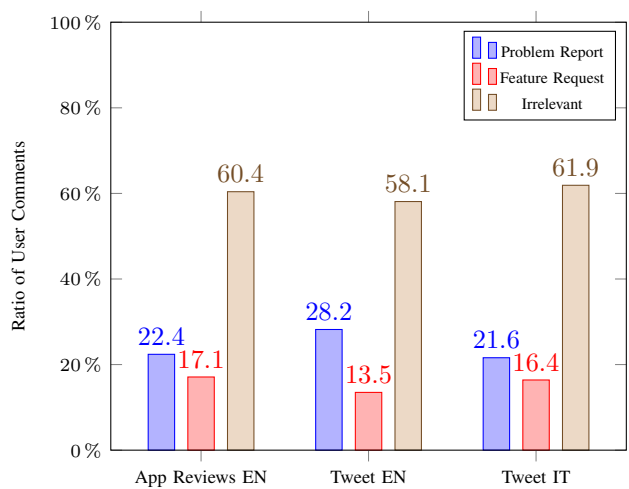

We evaluate our approaches on three different benchmark data sets (see Fig.~\ref{fig:data_sets}). The first data set consists of 6,406 English app reviews retrieved from the Google Play Store and Apple AppStore. The second data set includes 10,364 English tweets, while the third data set consists of 15,802 Italian tweets. All user comments have been annotated with respect to the following three classes:

\begin{enumerate}
    \item \textit{Problem report}: A user comment represents a problem report if it describes a concrete problem resp. bug in the app (e.g., \textit{Since the recent update I cannot upload new files to my account}). 
    \item \textit{Feature request}: A user comment represents a feature request when the user explicitly asks for new functionality (e.g., \textit{I would like to be able to assign members to groups}). User comments that request more information or suggest improvements to existing functionality are also included in this class. 
    \item \textit{Irrelevant}: A user comment is irrelevant to the development team if it neither expresses a problem report nor a feature request (e.g., \textit{I really love the app}).
\end{enumerate}

\subsection{Evaluation Procedure}\label{evalpro}

\textbf{Cross Validation} 
The performance of DL models depends heavily on the network architecture, as well as the hyperparameters used. Therefore, we compare the performance of our models using different hyperparameter configurations. To this end, we optimize the learning rate using the \textit{Tree-structured Parzen Estimator} (TPE) algorithm~\cite{bergstra11}. We achieve the best performance with a learning rate of 1e-05 and a batch size of 32. On account of the small size of our training set, we apply 3-fold cross-validation and train our model for 2 epochs on the training data. After this optimization, we evaluate our best performing model on the test set.

\textbf{Evaluation Metrics} We use standard metrics for evaluating our approaches: Accuracy, Precision, Recall, and F1 score. During the training process, we check the validation accuracy periodically in order to keep the model’s checkpoint with the best validation performance. When interpreting the metrics, it is important to consider which misclassification (False Negative or False Positive) matters most resp. causes the highest costs~\cite{Berry17}. We favor Recall over Precision because it is easier for users to discard False Positives than to manually detect False Negatives in the huge amount of user comments. In addition, a low Recall is associated with the risk of missing important information about urgent bugs or necessary features. Consequently, we seek high Recall to minimize the risk of missed RE-related user comments and acceptable Precision to ensure that users are not overwhelmed by False Positives. 

  \begin{figure*}
        \centering
         \fbox{\includegraphics[trim=2cm 3cm 1.5cm 4.5cm, width=\textwidth]{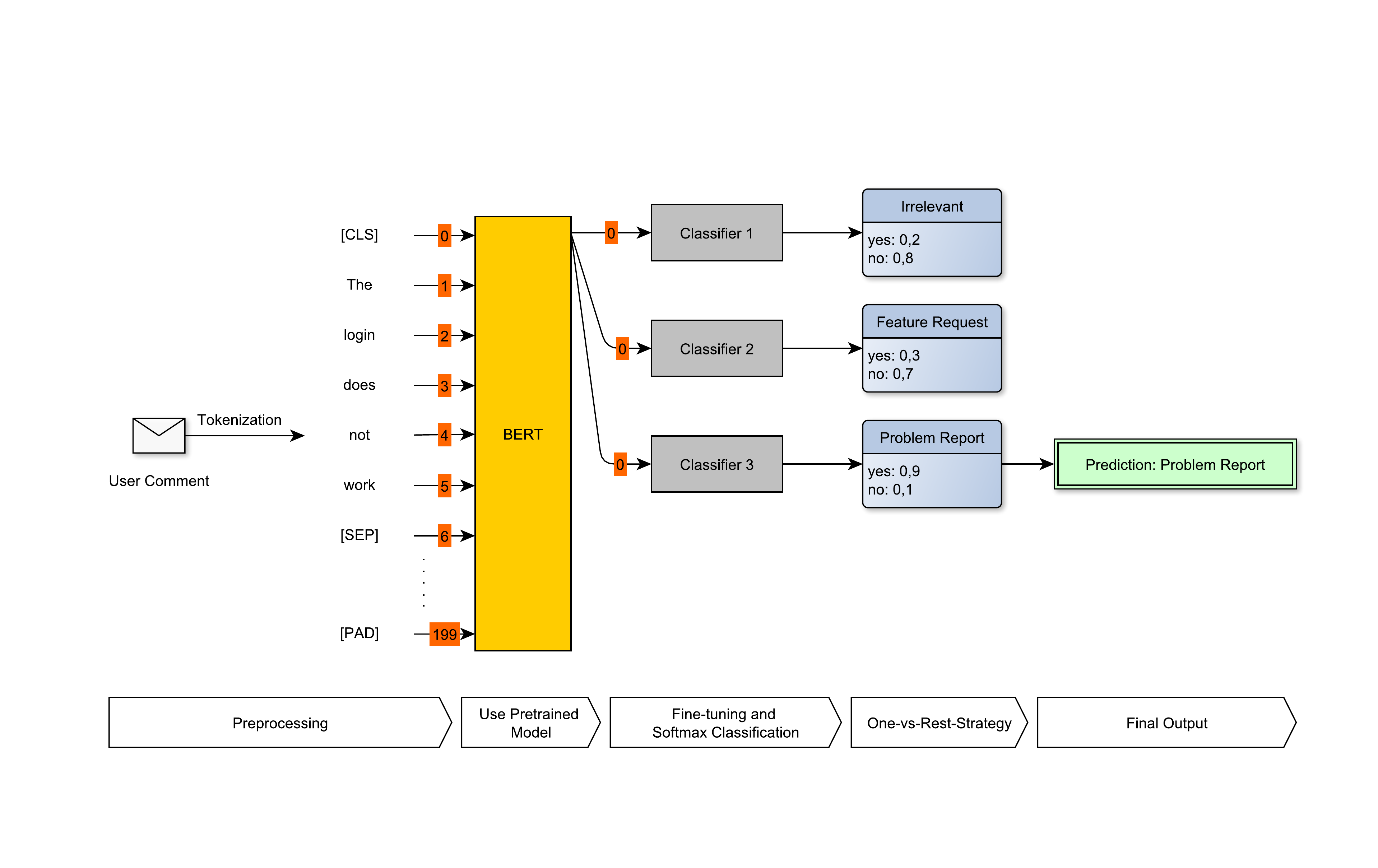}}
    \caption{Our \emph{BERT} pipeline for the prediction of user comments and app reviews. Each user comment is transformed into a \emph{BERT} embedding with a fixed length of 200 tokens. Only the CLS token (see token with id 0) is fed into the three softmax classifiers. In the present case, we achieve the highest probability for the problem report class.}\label{BERTclassification}
    \end{figure*}

\section{Transfer Learning for Classification of User Comments}
The Bidirectional Encoder Representations from Transformers (BERT) model~\cite{devlin19} is pre-trained on large corpora and can therefore easily be fine-tuned for any downstream task without the need for much training data (transfer learning). \emph{BERT} adapts the encoder-structure of the Transformer architecture by Vaswani et al.~\cite{vaswani2017attention} and is available in two variants~\cite{devlin19}. We utilize the \emph{BERT} base model which has twelve encoder-layers, uses 768 hidden units, and twelve attention heads. In our paper, we make use of the fine-tuning mechanism of \emph{BERT} and investigate to which extent it can be used for the classification of user comments in our particular domain. 

\textbf{Preprocessing}
Since we are dealing with Tweets, many of our user comments contain mentions to Twitter accounts preceded by an ``@'' symbol. For example: ``Dear, @MyCompany, my WiFi is down''. In order to avoid these mentions being split into different tokens by the \emph{BERT} tokenizer, we first replace them with the word @mention, and then add this word as a token to the \emph{BERT} tokenizer.
After this, we tokenize each comment. \emph{BERT} requires an input sequence with a fixed length (maximum 512 tokens). Therefore, for comments that are shorter than this fixed length, padding tokens (PAD) are inserted to adjust all comments to the same length. To choose a suitable fixed length, we analyzed the lengths of the user comments in our data sets. Even with a fixed length of 200 tokens, we cover more than 99~\% of the user comments. User comments containing more tokens are shortened accordingly. Since this is only a small amount, only little information is lost. Thus, we chose a fixed length of 200 tokens instead of the maximum possible 512 tokens to keep \emph{BERT's} computational requirements to a minimum (see Fig.~\ref{BERTclassification}). Other special tokens, such as the classification (CLS) token, are also inserted in order to provide further information of the comment to the model. CLS is the first token in the sequence and represents the whole comment (i.e., it is the pooled output of all tokens of a comment). For our classification task, we mainly use this token because it stores the information of the whole comment. We feed the pooled information into a single-layer feedforward neural network that uses a softmax layer, which calculates the class probabilities. In order to reduce overfitting, we also integrate a dropout layer (dropout probability = 0.3) in our pipeline.

\textbf{Binary Relevance} We are facing a multi-class classification problem since we want to classify a user comment into one of the three described categories. Similar to Stanik et al.\cite{stanik19}, we apply \textit{Binary Relevance} to solve the multi-class classification problem. Specifically, we decompose the multi-label task into three independent binary classification tasks (one per class label). As a result, we train three different softmax classifiers: one model predicts whether the user comment represents a problem report or not. A second model predicts whether a certain user comment describes a feature request or not. Finally, a third model predicts whether the comment is irrelevant or not. Each model predicts a class membership probability. We use the argmax of these scores as the final prediction (see Fig.~\ref{BERTclassification}).

\textbf{Monolingual vs. Multilingual Prediction}
As described in Section~\ref{study_objects}, our data set contains user comments in both English and Italian. A simple solution to process both languages is to train a separate model for each language and use it for the classification of user comments (\textit{monolingual prediction}). Alternatively, a multilingual model can be trained, which is capable of handling multiple languages (\textit{multilingual prediction}). Multilingual prediction is clearly a more difficult task than monolingual prediction because the model must generalize across different languages (cross-lingual transfer learning). However, recent studies achieved promising results by applying multilingual models~\cite{pires-etal-2019-multilingual,muller20,k2020crosslingual}. In our experiments, we investigate the multilingualism of \emph{BERT} in the classification of user comments. To this end, we train the Multilingual \emph{BERT} model (M-BERT) on the two English data sets and evaluate it on the Italian data set and vice versa. \keys{M-BERT} is pre-trained on the concatenation of Wikipedia corpora from 104 languages and follows the same Transformer architecture as the \emph{BERT} base model. To make the results of \keys{M-BERT} comparable to the results of the monolingual models, we use the same test sets. Table~\ref{modeloverview} provides an overview of the configuration of our two monolingual models and the multilingual model.

\textbf{Class Imbalance}
All three data sets are unbalanced (see Fig.~\ref{fig:data_sets}). To avoid bias of the model towards the irrelevant class, we used \textit{Random Undersampling}. We randomly select user comments from the majority class and exclude them from the data set until a balanced distribution is achieved. In our experiments, we study the impact of the balancing strategy on the performance of \keys{M-BERT} and the monolingual \emph{BERT} models.

\begin{table*}[]
\caption{Overview of trained monolingual and multilingual models}
\label{modeloverview}
\resizebox{\textwidth}{!}{\begin{tabular}{@{}llll@{}}
\toprule
\textbf{Model name} & \textbf{Model id}            & \textbf{Configuration}                           & \textbf{Pre-training}                                                       \\ \midrule
\rowcolor[HTML]{EFEFEF} 
\keys{English-BERT}        & bert-base-uncased              & 12-layer, 768-hidden, 12-heads, 109M parameters. & Trained on uncased English text retrieved from BookCorpus and Wikipedia.      \\
\keys{Italian-BERT}       &              dbmdz/bert-base-italian-cased                &             12-layer, 768-hidden, 12-heads, 109M parameters               &                         Trained on cased Italian text retrieved from OPUS corpora and Wikipedia.                                                      \\
\rowcolor[HTML]{EFEFEF} 
\keys{M-BERT}              & bert-base-multilingual-uncased & 12-layer, 768-hidden, 12-heads, 179M parameters. & Trained on uncased text retrieved from concatenated Wikipedia data for 104 languages. \\ \bottomrule
\end{tabular}}
\end{table*}

\section{Experimental Results}
This section compares the monolingual and multilingual \emph{BERT} models with state-of-the-art methods for user comment classification. In addition, we study the impact of \textit{Random Undersampling} on the performance of our models and investigate whether the multilingual models can achieve similar results as the monolingual models.

\textbf{Prediction of English App Reviews}
We found that \keys{English-BERT} outperforms the state-of-the-art methods independent of the used balancing method. Across all three classes, it shows better predictive power than the baseline systems and increases the F1 score by 5\%. Especially \keys{English-BERT} with under-sampling demonstrates very good Recall values of 90\% for problem reports and 87\% for feature requests. Across all three categories, it improves the state-of-the-art by 10\% in Recall. In addition, the model achieves good Precision values of 77\% for problem reports and 75\% for feature requests, avoiding that users are overwhelmed with False Positives. \keys{M-BERT} does not perform well in the prediction of English app reviews. Regardless of the application of \textit{Random Undersampling}, the model shows a bias towards the irrelevant class and tends to predict almost all user comments as irrelevant. Across all classes, \keys{M-BERT} achieves a maximum F1 score of 47\% and thus shows a substantially lower predictive power than the state-of-the-art methods.

\textbf{Prediction of English Tweets}
Similar to the prediction of the English app reviews, the monolingual models show better predictive power on English tweets than the existing approaches. Independent of the used balancing method, \keys{English-BERT} increases the macro-F1 scores by 5\%. However, the Recall scores can only be marginally increased. For example, \keys{English-BERT} with under-sampling shows an increase of 4\% for the irrelevant class and 4\% for the problem reports. For the feature requests, however, it performs worse (difference 3\%). The monolingual models achieve a slight improvement with regard to the Precision values, which ultimately results in an increase in the F1 scores. \keys{English-BERT} without under-sampling increases the Precision value of problem reports and the irrelevant class by 8\% each. Only the Precision in the case of feature requests decreases by 2\%. Similar to the prediction of English app reviews, \keys{M-Bert} does not perform well in the prediction of English tweets. We still observe the bias towards the irrelevant class, which prevents the model from being used in practice. Independent of the used balancing method, \keys{M-Bert} achieves a poorer macro-F1 score than the state-of-the-art methods.

\textbf{Prediction of Italian Tweets}
Table~\ref{experimentalresults} reveals that \keys{Italian-BERT} with under-sampling outperforms the state-of-the-art methods. It shows significantly better predictive power by achieving a higher F1 score across all three categories (difference of 8\% to traditional ML methods). The model improves the Recall values for the irrelevant class by 4\% and for feature requests by 3\%, but is not able to beat the performance of the traditional methods for problem reports (difference of 4\%). Substantial differences emerge, however, with respect to the Precision for all three categories. For example, for a feature request, \keys{Italian-BERT} with under-sampling was able to increase the Precision value by 18\%. \keys{Italian-BERT} without under-sampling demonstrates a similar good performance. However, the model achieves a worse Recall across all categories, making \keys{Italian-BERT} with under-sampling the best model for our use case. Both multilingual models perform worse compared to the monolingual models. Interestingly, \keys{M-BERT} with under-sampling achieves similar results to the DL methods, but cannot beat the performance of the traditional ML methods. 

\textbf{Monolingual vs. Multilingual Prediction}
Across all three data sets, the monolingual models perform significantly better than the multilingual models. In our case, we cannot confirm the finding of other studies that multilingual models can achieve similar results to monolingual models~\cite{wu-dredze-2019-beto,conneau2020unsupervised}. A comparison between the two trained multilingual models reveals interesting results. \keys{M-BERT} shows significantly better performance when it is trained on English data and then applied to the Italian tweets, while the cross-lingual transfer learning from Italian to English did not perform well. For both the prediction of App Reviews and English Tweets, \keys{M-BERT} shows a clear bias towards the irrelevant class. Generally, we would expect such a performance discrepancy to be caused by unevenly distributed training data. This assumption can be rejected in our case, as our benchmark data set contains a similar amount of user comments from both languages: 16,770 English user comments and 15,802 Italian user comments. Our finding that the cross-lingual transfer from English to Italian is more effective than from Italian to English corroborates the findings of other related studies. For example, Pires et al.~\cite{pires-etal-2019-multilingual} found that \keys{M-BERT} performs better when it is fine-tuned for English and then used for Named Entity Recognition and POS tagging on a German, Spanish, Dutch or Italian data set than vice versa.

\textbf{Impact of Random Undersampling}
In our experiment, applying \textit{Random Undersampling} did not necessarily lead to a performance gain. We observed that the balancing strategy had the greatest impact on the performance of \keys{M-BERT}, which was trained on the English data set. \textit{Random Undersampling} increased the F1 scores for problem reports (difference 3\%) and feature requests (difference 14\%). Across all categories, the F1 score increased by 6\%, which underlines the improved predictive power of the model. In the case of \keys{M-BERT}, which was trained on the Italian data set, the predictive power also increased. However, despite \textit{Random Undersampling}, the bias of the model with respect to the irrelevant class could not be prevented. We found no significant impact of \textit{Random Undersampling} on the performance of the monolingual models. Especially in the case of \keys{English-BERT} the predictive power did not change considerably.

\begin{tcolorbox}[breakable, enhanced jigsaw]
\textbf{Summary of Experiments:} \keys{English-BERT} and \keys{Italian-BERT} achieve new state-of-the-art results for the classification of all three benchmark data sets. However, with respect to Recall as our main evaluation criterion, the monolingual models do not yield a large performance gain compared to traditional ML methods in the prediction of English tweets. \keys{M-BERT} does not perform well on any of the three benchmark data sets and performs worse than the state-of-the-art methods.
\end{tcolorbox}

\begin{table*}[]
\caption{Experimental results. Recall values of at least 90\% are marked in \textbf{bold}. We underline the best metric values over all classes for each benchmark data set. Accuracy values of the state-of-the-art methods have not been reported by Stanik et al.~\cite{stanik19}}.
\label{experimentalresults}
\resizebox{\textwidth}{!}{\begin{tabular}{ll|llll|llll|llll}
\hline
                                                                                                                                               &                 & \multicolumn{4}{c|}{\textbf{App Review EN}} & \multicolumn{4}{c|}{\textbf{Tweet EN}} & \multicolumn{4}{c}{\textbf{Tweet IT}} \\ \hline
                                                                                                                                               &                 & acc   & pre   & rec            & f1         & acc   & pre   & rec            & f1    & acc   & pre   & rec            & f1   \\
\rowcolor[HTML]{EFEFEF} 
\cellcolor[HTML]{EFEFEF}                                                                                                                       & irrelevant      & -     & 0.88  & 0.89           & 0.89       & -     & 0.73  & 0.75           & 0.74  & -     & 0.78  & 0.89           & 0.83 \\
\rowcolor[HTML]{EFEFEF} 
\cellcolor[HTML]{EFEFEF}                                                                                                                       & problem report  & -     & 0.83  & 0.75           & 0.79       & -     & 0.46  & 0.82           & 0.59  & -     & 0.51  & 0.88           & 0.65 \\
                                  \rowcolor[HTML]{EFEFEF} 
\cellcolor[HTML]{EFEFEF}                            & feature request & -     & 0.68  & 0.76           & 0.72       & -     & 0.32  & 0.7            & 0.43  & -     & 0.47  & 0.82           & 0.6  \\
                                                            \rowcolor[HTML]{EFEFEF} 
\multirow{-4}{*}{\cellcolor[HTML]{EFEFEF}Traditional\_ML\_methods~\cite{stanik19}}   
& all classes (avg.) & -     & 0.79  & 0.8         & 0.8       & -     & 0.5  & 0.75          & 0.58  & -     & 0.58  & 0.86           & 0.69  \\
                                                                                                                                               & irrelevant      & -     & 0.78  & \textbf{0.93}  & 0.85       & -     & 0.74  & 0.7            & 0.72  & -     & 0.85  & 0.77           & 0.81 \\
                                                                                                                                               & problem report  & -     & 0.46  & 0.6            & 0.52       & -     & 0.51  & 0.42           & 0.46  & -     & 0.62  & 0.57           & 0.59 \\
      & feature request & -     & 0.69  & 0.79           & 0.74       & -     & 0.51  & 0.57           & 0.54  & -     & 0.47  & 0.82           & 0.6  \\ \multirow{-4}{*}{DL\_methods~\cite{stanik19}}                                             & all classes (avg.) & -     & 0.64  & 0.77          & 0.7  & -     & 0.58  & 0.56           & 0.57  & -     & 0.64  & 0.72           & 0.66  \\\hline
\rowcolor[HTML]{EFEFEF} 
\cellcolor[HTML]{EFEFEF}                                                                                                                       & irrelevant      & 0.93  & 0.94  & \textbf{0.95}  & 0.94       & 0.74  & 0.76  & 0.79           & 0.78  &       &       & \textbf{}      &      \\
\rowcolor[HTML]{EFEFEF} 
\cellcolor[HTML]{EFEFEF}                                                                                                                       & problem report  & 0.91  & 0.77  & \textbf{0.9}   & 0.83       & 0.68  & 0.46  & 0.86           & 0.6   &       &       &                &      \\
\rowcolor[HTML]{EFEFEF} 
\cellcolor[HTML]{EFEFEF}                                                  & feature request & 0.93  & 0.75  & 0.87           & 0.8        & 0.83  & 0.42  & 0.67           & 0.52  &       &       &                &      \\
\rowcolor[HTML]{EFEFEF} 
\multirow{-4}{*}{\cellcolor[HTML]{EFEFEF}\keys{English-BERT}\_with\_undersampling}                                                       & all classes (avg.) & \underline{0.92} & 0.82  & \underline{0.90}          & \underline{0.85}        & 0.75  & 0.54  & \underline{0.77}           & \underline{0.63}  &       &       &                &      \\
                                                                                                                                               & irrelevant      & 0.92  & 0.97  & \textbf{0.9}   & 0.93       & 0.75  & 0.82  & 0.72           & 0.77  &       &       &                &      \\
                                                                                                                                               & problem report  & 0.91  & 0.78  & 0.87           & 0.82       & 0.77  & 0.59  & 0.63           & 0.61  &       &       &                &      \\         & feature request & 0.93  & 0.8   & 0.81           & 0.8        & 0.86  & 0.49  & 0.55           & 0.52  &       &       &                &      \\
\multirow{-4}{*}{\keys{English-BERT}\_no\_undersampling}                                                                                 & all classes (avg.) & \underline{0.92}  & \underline{0.85}  & 0.86           & \underline{0.85}        & \underline{0.79}  & \underline{0.63}  & 0.63           & \underline{0.63}  &       &       &                &      \\
\rowcolor[HTML]{EFEFEF} 
\cellcolor[HTML]{EFEFEF}                                                                                                                       & irrelevant      &       &       & \textbf{}      &            &       &       &                &       & 0.88  & 0.88  & \textbf{0.93}  & 0.9  \\
\rowcolor[HTML]{EFEFEF} 
\cellcolor[HTML]{EFEFEF}                                                                                                                       & problem report  &       &       & \textbf{}      & \textbf{}  &       &       &                &       & 0.84  & 0.59  & 0.84           & 0.69 \\
\rowcolor[HTML]{EFEFEF} 
\cellcolor[HTML]{EFEFEF}                                                                                            & feature request &       &       &                &            &       &       &                &       & 0.9   & 0.65  & 0.85           & 0.73 \\
\rowcolor[HTML]{EFEFEF} 
\multirow{-4}{*}{\cellcolor[HTML]{EFEFEF}\keys{Italian-BERT}\_with\_undersampling}                                                        & all classes (avg.) &       &       &                &            &       &       &                &       & \underline{0.87}   & 0.70  & \underline{0.87}          & \underline{0.77} \\
                                                                                                                                               & irrelevant      &       &       & \textbf{}      &            &       &       &                &       & 0.87  & 0.91  & 0.87           & 0.89 \\
                                                                                                                                               & problem report  &       &       &                &            &       &       &                &       & 0.88  & 0.7   & 0.76           & 0.73 \\  & feature request &       &       &                &            &       &       &                &       & 0.84  & 0.74  & 0.62            & 0.68 \\
\multirow{-4}{*}{\keys{Italian-BERT}\_no\_undersampling}                                                                                  & all classes (avg.) &       &       &                &            &       &       &                &       & 0.86  & \underline{0.78}  & 0.75            & 0.76 \\
\rowcolor[HTML]{EFEFEF} 
\cellcolor[HTML]{EFEFEF}                                                                                                                       & irrelevant      & 0.6   & 0.6   & \textbf{1}     & 0.75       & 0.64  & 0.63  & \textbf{0.95}  & 0.75  &       &       &                &      \\
\rowcolor[HTML]{EFEFEF} 
\cellcolor[HTML]{EFEFEF}                                                                                                                       & problem report  & 0.82  & 0.69  & 0.41           & 0.51       & 0.73  & 0.56  & 0.3            & 0.39  &       &       &                &      \\
\rowcolor[HTML]{EFEFEF} 
\cellcolor[HTML]{EFEFEF}  & feature request & 0.84  & 0.55  & 0.1            & 0.17       & 0.85  & 0.46  & 0.53           & 0.49  &       &       &                &      \\
\rowcolor[HTML]{EFEFEF} 
\multirow{-4}{*}{\cellcolor[HTML]{EFEFEF}\begin{tabular}[c]{@{}l@{}}\keys{M-BERT}\_with\_undersampling \\ (fine-tuned on Italian data)\end{tabular}} & all classes (avg.) & 0.75  & 0.61  & 0.5            & 0.47       & 0.74  & 0.55  & 0.59           & 0.54  &       &       &                &      \\
                                                                                                                                               & irrelevant      & 0.64  & 0.63  & \textbf{0.99}  & 0.77       & 0.59  & 0.59  & \textbf{0.99}  & 0.74  &       &       &                &      \\
                                                                                                                                               & problem report  & 0.78  & 0.78  & 0.09           & 0.17       & 0.73  & 0.62  & 0.1            & 0.17  &       &       &                &      \\
                     & feature request & 0.83  & 0.5   & 0.01           & 0.02       & 0.87  & 0.69  & 0.13           & 0.22  &       &       &                &      \\
\multirow{-4}{*}{\begin{tabular}[c]{@{}l@{}}\keys{M-BERT}\_no\_undersampling \\ (fine-tuned on Italian data)\end{tabular}}                           & all classes (avg.) & 0.75  & 0.63   & 0.36           & 0.32       & 0.73  & \underline{0.63}  & 0.40          & 0.37  &       &       &                &      \\
\rowcolor[HTML]{EFEFEF} 
\cellcolor[HTML]{EFEFEF}                                                                                                                       & irrelevant      &       &       &                &            &       &       &                &       & 0.72  & 0.71  & \textbf{0.93}  & 0.81 \\
\rowcolor[HTML]{EFEFEF} 
\cellcolor[HTML]{EFEFEF}                                                                                                                       & problem report  &       &       &                &            &       &       &                &       & 0.82  & 0.64  & 0.43           & 0.52 \\
\rowcolor[HTML]{EFEFEF} 
\cellcolor[HTML]{EFEFEF}    & feature request &       &       &                &            &       &       &                &       & 0.78  & 0.41  & 0.82           & 0.55 \\
\rowcolor[HTML]{EFEFEF} 
\multirow{-4}{*}{\cellcolor[HTML]{EFEFEF}\begin{tabular}[c]{@{}l@{}}\keys{M-BERT}\_with\_undersampling \\ (fine-tuned on English data)\end{tabular}} & all classes (avg.) &       &       &                &            &       &       &                &       & 0.77  & 0.58  & 0.72           & 0.62 \\
                                                                                                                                               & irrelevant      &       &       &                &            &       &       &                &       & 0.78  & 0.78  & \textbf{0.9}   & 0.83 \\
                                                                                                                                               & problem report  &       &       &                &            &       &       &                &       & 0.82  & 0.64  & 0.39           & 0.49 \\                    & feature request &       &       &                &            &       &       &                &       & 0.86  & 0.65  & 0.27           & 0.38 \\
\multirow{-4}{*}{\begin{tabular}[c]{@{}l@{}}\keys{M-BERT}\_no\_undersampling \\ (fine-tuned on English data)\end{tabular}}                           & all classes (avg.) &       &       &                &            &       &       &                &       & 0.82  & 0.69  & 0.52           & 0.56 \\\hline
\end{tabular}}
\end{table*}

\section{Threats to Validity}
This section discusses potential threats to validity of our research methodology and experimental results.

\textbf{Internal Validity}
A potential threat to internal validity arises from the fact that the app review set was labeled by students. The coders were briefed before the annotation process and only peer-agreed labels were included in the data set~\cite{Maalej2017}. However, it is questionable to what extent students can represent development teams. Do they have a similar understanding of which user comments, for example, already require new features and which still express bugs in existing functionalities? We used the data sets as they are well-known and accepted in the community and let us directly compare our results to the state-of-the-art approaches.

\textbf{External Validity}
To test the generalization performance of our models on unseen data we applied cross validation, which is widely used in research and practice. However, a potential threat to the generalization of our results might be that the data sets are not representative for user comments in app stores and social media in general. For example, our results do not allow any conclusion about the applicability of the models to user comments from other sources (e.g., Instagram). Nevertheless, we assume that the models can be applied in this case as well, because user comments usually follow a similar language style and structure. Validation of this claim requires further empirical investigation.

\textbf{Construct Validity}
To mitigate potential risks to construct validity, we applied widely used experimental designs and evaluation metrics. In order to render our approaches comparable to the state-of-the-art methods, we followed the same research methodology as Stanik et al.~\cite{stanik19}. Consequently, we also trained one classifier per class and applied 3-fold cross-validation on the training set for each hyperparameter configuration. 

\section{Transfer Learning - the silver bullet?}
As a result of the continuous progress in the NLP community, we witness new models being introduced almost every month. Researchers working with natural language are currently less concerned with the question ``is there an NLP model I can use for my task?'' and more with the question ``which NLP model do I use?''. Consequently, simple rule-based approaches or shallow ML methods are often overlooked and one tends to directly apply heavyweight DL models. Previous studies have already shown that these models do not necessarily lead to better performance~\cite{FakhouryANKA18,menzies17}. In our setting, even the use of pre-trained language models with several million parameters does not necessarily lead to a large performance boost. For example, our monolingual models show similar performance in the prediction of English tweets as traditional ML methods operating on simple TF-IDF word embeddings. In this case, we observe that carefully elicited features by domain experts are able to compete with the 109M automatically obtained parameters. Our experiments also indicate that cross-lingual transfer learning is unable to compete with existing baseline systems for our use case. So far, other studies on cross-lingual transfer learning have mainly achieved good results for token-level prediction (e.g., POS tagging~\cite{pires-etal-2019-multilingual}). However, our task is substantially more complex since the prediction is done at sentence level. The multilingual model must therefore not only generalize at word level across multiple languages but must even be able to understand the semantics of full sentences. In our experiments, the cross-lingual transfer across sentences seems to work to a certain degree - especially when transferring from English to Italian. However, the multilingual models do not perform in a way that allows for practical use (cf. Recall and Precision condition in Section~\ref{evalpro}). A similar observation is made by Artetxe et al.~\cite{artetxe2019} who use multilingual models for Natural Language Inference (NLI), where the prediction is also conducted on sentence level. We hypothesize that it still requires more research to increase the robustness of multilingual models and make them usable for practitioners.

\section{Related Work}
There is a large body of work on the identification of RE-related information in user comments. Maalej et al.~\cite{Maalej2017} use traditional machine learning to categorize English app reviews into bug reports and feature requests. A similar approach is presented by Guzman et al.~\cite{Guzman15}, who classify app reviews into even more detailed categories: bug report, feature strength, feature shortcoming, user request, praise, complaint, and usage scenario. Chen et al.~\cite{chen14} use topic modeling to group related user comments and present the \emph{AR miner} which can be used to visualize the most ''informative`` user comments for development teams. Dhinakaran et al.~\cite{Dhinakaran18} show how active learning can be used to significantly reduce the human effort required for training app review classifiers. Automatically mining feature requests and bug reports from tweets was first addressed by Guzman et al.~\cite{guzman17}, and Williams and Mahmoud~\cite{williams17}. Both use traditional machine learning to detect feature requests and bug reports in tweets. Guzman et al.~\cite{guzman17} also present how to map the relevance of tweets by using different ranking methods. The most recent work on our use case originates from Stanik et al.~\cite{stanik19} and Vliet et al.~\cite{vliet20}. Stanik et al.~\cite{stanik19} compare traditional ML methods with DL methods and show that both methods perform similarly in the classification of App Reviews and Tweets. Vliet et al.~\cite{vliet20} present a crowdsourcing-based approach for the classification of online feedback. Specifically, they train crowd workers to perform the classification. 

\textbf{Relation to Our Work} In contrast to previous work, we are the first to investigate the potential of transfer learning methods for the classification of user comments. The novelty of our work stems from the fact that we use both monolingual and multilingual models and study whether the cross-lingual generalization ability of multilingual TL models is applicable to reliably classify user comments written in different languages.

\section{Conclusion}
Identifying RE-related information in user comments is a challenging task due to the following reasons: (1) about 70\% of the user comments contain noisy, irrelevant information, (2) the amount of user comments grows daily making manual analysis unfeasible, and (3) user comments on globally distributed applications are written in different languages. Existing approaches focus on traditional ML and DL methods and fail to classify user comments with reasonable performance (i.e., they do not achieve high Recall and acceptable Precision which is necessary for our use case). Recent studies found that TL is capable of optimizing state-of-the-results for a number of RE use cases (e.g., requirements classification~\cite{Hey20}). However, we still lack knowledge about the potential of TL methods for user comment classification. This paper addresses the research gap and compares monolingual and multilingual TL models with state-of-the-art methods for this task. We demonstrate that monolingual \emph{BERT} models achieve new state-of-art results in the classification of all three benchmark data sets. However, we also highlight that TL models do not necessarily lead to a huge performance gain. In fact, we found no large difference between the performance of \emph{BERT} models and traditional ML methods in the classification of English Tweets. We also show that so far cross-lingual transfer learning is not able to keep up with the performance of the existing baseline systems for user comment classification.

\section*{Acknowledgements}
We would like to thank Walid Maalej for sharing the benchmark data sets.
In addition, we would like to acknowledge that this work was in parts supported by the KKS foundation through the S.E.R.T. Research Profile project at Blekinge Institute of Technology.

\bibliographystyle{IEEEtran}
\bibliography{references}

\end{document}